\documentclass{article}

     \usepackage[final]{neurips_2019}

\usepackage[utf8]{inputenc} 
\usepackage[T1]{fontenc}    
\usepackage{hyperref}       
\usepackage{url}            
\usepackage{booktabs}       
\usepackage{amsfonts}       
\usepackage{nicefrac}       
\usepackage{microtype}      
\usepackage[pdftex]{graphicx}

\title{Explainable Artificial Intelligence (XAI) for Increasing User Trust in Deep Reinforcement Learning Driven Autonomous Systems}

\author{%
  Jeff Druce\\
  Charles River Analytics\\
  Cambridge, MA \\
  \texttt{jdruce@cra.com} \\
  \And
  Michael Harradon \\
  Charles River Analytics\\
  Cambridge, MA \\
  \texttt{mharradon@cra.com} \\
  \AND
  James Tittle \\
  Charles River Analytics\\
  Cambridge, MA \\
  \texttt{jtittle@cra.com} \\
}

\begin{document}

\maketitle

\begin{abstract}
We consider the problem of providing users of deep Reinforcement Learning (RL) based systems with a better understanding of when their output can be trusted. We offer an explainable artificial intelligence (XAI) framework that provides a three-fold explanation: a graphical depiction of the systems generalization and performance in the current game state, how well the agent would play in semantically similar environments, and a narrative explanation of what the graphical information implies. We created a user-interface for our XAI framework and evaluated its efficacy via a human-user experiment. The results demonstrate a statistically significant increase in user trust and acceptance of the AI system with explanation, versus the AI system without explanation. 
\end{abstract}

\section{Introduction}

Artificial intelligence is playing an increasingly prominent role in societies throughout the world. The recent explosion in their popularity can arguably be contributed to their dramatic increase in performance afforded by deep learning (DL) architectures \cite{deng2014tutorial}, and associated hardware advances \cite{schmidhuber2015deep}. E.g., computer vision has benefited from Convolutional Neural Networks (CNNs) \cite{lecun1995convolutional}, resulting in near human level perception for object identification, tracking, and segmentation. Natural language processing has benefited from recurrent neural network architectures, increasing performance in tasks such as machine translation \cite{devlin2018bert}. Autonomous agents, such as those controlled by deep RL policy networks, have been of particular importance. Deep RL architectures as function approximators have demonstrated super-human performance in a variety of Atari video game environments \cite{mnih2015human}, and are showing promise in arenas of high complexity such as StarCraft 2 \cite{alphastarblog}. This high-performing autonomy motivates its widespread deployment in a variety of applications. 

However, a drawback of these high-performing neural network architectures, and hence deep RL based systems, is that they are inherently black-box and offer little insight into \emph{why} they produce the output they do \cite{gunning2017explainable}. Deep learning architectures consist of a (typically large) number of cascading mathematical operations mapping input to output. It is this large number of operations that allows modeling complex, non-linear patterns in data. As a consequence of these large number of operations, the aspects of the input that were inflectional in formulating system output become blurred. In control networks, where the output could govern the actions of highly impactful systems (e.g., autonomous vehicles), the lack of explainability is worrisome. Even more problematic, deep RL based systems have been shown to be highly dependent on the exact structure of the input, producing wildly different output with only small perturbations to the input  \cite{generalization}. Users employing these system can be left with agents producing erratic behavior with little indication as to why. 

Explainable artificial intelligence (XAI) methods offer means to peer inside the black-box of deep RL based systems. Many XAI methods have been developed to offer explanations for DL systems including: saliency based-method \cite{selvaraju2017grad}, local interpretable model approximations \cite{ribeiro2016should}, and activation propagation methods \cite{bach2015pixel}. These methods can be effective for providing users with information as to what features in data-space were most influential in formulating system output (e.g., an attention map highlighting the ears of a husky in a dog-cat classification task). These methods are effective in helping explain \emph{why} black-box systems produce the output they do, but don't necessarily foster trust in the AI system. 

\section{Approach}

In this work, we are concerned with engendering user trust in deep RL based autonomous agents. Our domain of interest is the Atari game, Amidar. In this game, a player navigates through a board collecting points for traversing horizontal and vertical line segments, while avoiding enemies to prevent being killed. The q-network we are considering, is a convolutional neural network consistent with the architecture in \cite{mnih2015human}, where the output is the expected total discounted future reward for each action. An experience memory mechanism is employed to mitigate the problems associated with correlated data and non-stationary distributions. 

The interest in conveying trustworthiness versus explaining low level actions was motivated by the observation that our autonomous agent only performed well in an environment that was virtually identical to what it had seen in training. Specifically, we observed our q-network policy controlling the agent produced extremely erratic behavior when small semantically meaningful interventions to game environment were performed. E.g., when we removed a single enemy from the game, the agent performed significantly worse on average. This behavior was magnified the more enemies were removed. This implied the agent had not learned a robust strategy for the game, but was simply using the enemies as a "signal" as to how to act; a more detailed analysis is available in \cite{generalization}. 

With this potential brittleness in mind, in order to build trust in a deep RL based system, we argue that not only information on how well the system is currently performing is necessary, but also how well it is generalizing away from its training set. Therefore, we consider two key metrics: 1) the state Value Estimate Error (VEE); and 2) the distance to the nearest training sample (DNTS). The output of the underlying deep q-network responsible for the system output is the total discounted future reward for each available action. Thus, we can convey the performance of the system by comparing its output against the actual reward accumulated by the agent. I.e., a low VEE implies the agent has an understanding of the current state value, and a high error means it erroneously estimated the value of the state. DNTS provides a notion of how similar the DL \emph{perceives} the current environment is relative to a sample from training \cite{mandelbaum2017distance}. A low-distance implies the agent considers this as something similar to what it has seen in training, and a large distance corresponds to something not similar to what it has seen in training. Therefore, the DNTS metric gives the user a sense of how well the system is generalizing away from its training set. More details on these metrics are provided in Section \ref{approach}. 

We evaluate the effectiveness of our XAI system with an 18-subject user study. Subjects were broken into two groups, those with our XAI system, and those without. Those without the interface showed a statistically significant decrease in user trust and acceptance of the AI system. As an additional study, we gave questions which tested the users mental model of the AI. We presented this as a prediction task; the results of this case were inconclusive. 

\section{Conveying Trustworthiness} \label{approach}
At the core of approach is conveying information to a user as to whether they can trust the actions of an agent's underlying policy network. To convey trustworthiness, we explore how the underlying $Q$ network is performing in its current environment, how well the agent performs in a variety of scenarios similar to the current environment, and present an analysis to the user. 

\subsection{Metrics} \label{metrics}

Deep RL q-networks are typically function approximators, consisting of deep neural networks, where the function they are approximating estimates the value of the current state. Specifically, they act as control policies which can be greedily used (although stochasticity is often incorporated, especially in training to foster exploration) to produce agents that gain as much reward as possible. The total discounted reward, denoted by $Q^*$, and can be defined recursively as:

\begin{equation}
Q^*(s,a) = r(s,a) + \gamma \max_{a'} Q^*(s_{t+1},a')
\end{equation}

where $r$ is the reward, and $s$ is the state, and $a$ is the action. The deep network defining the $Q$ function is trained (via stochastic gradient decent) to minimize the error between the estimated total discounted future reward, and the actual reward. Therefore, to convey a notion of the underlying network's performance, we chose to provide information on how well the agent estimates its state (versus, e.g., the total points achieved, or any number of other performance metrics \cite{poole2010artificial})-- which we refer to as Value Estimate Error (VEE). Mathematically, this can expressed as

\begin{equation}
 VEE = \sum_{t=N}^{\infty} |{Q^*(s_t,a)} - V(s_t)|, 
\end{equation}

where $t$ indicates time, $N$ indicates the time index for the current game sate, $s$ is the state, $a$ is the action, and $V$ is the actual discounted future reward \footnote{We note that this requires "running forward" the episode until completion along the course of the episode. This, of course, cannot be done in real time. However, we note that the current cumulative discounted reward could be calculated, and used instead. This is a matter we intend to explore in future work.}. When the value of this comparison variable is high, the error in the policy network’s estimation implies it does not comprehend its environment well; when the value is low, it tends to understand its environment. Therefore, this variable provides an indication of how much the user can trust the agent in the current environment. The value for an example episode is plotted on the y-axis of the graph in Fig. \ref{UI}.  

The VEE by itself is not completely telling as to how well the agent is trustworthy. For example, if the VEE is low, but the current game state is something the agent has seen many time in training, the agent may not be robust. Conversely, if the current game state is extremely dissimilar from what the agent has seen in training, and the VEE is high, it does not necessarily imply the agent is not somewhat generalized. 

To introduce a notion of how well the underlying policy network has generalized, we introduce the Distance to Nearest Training Sample (DNTS) metric. Rather than do a comparison of the current game state to states directly in data space (i.e., the RBG image tensors), we calculate the $\ell_2$ distance of the feature embedding in the penultimate layer (the second to last layer). This is motivated by the fact that distances in the original domain would not reflect the \emph{semantic} difference in the data. E.g., for two game states, if the board games were identical (same ladders filled, and enemy locations), except for the agent location, the $\ell_2$ distance between them would be the same for any agent location. The feature embedding layer however, is effectively the "featurized" version of the data according to the q-network. Therefore, looking at the distance in this space, allows a comparison of the q-networks perception of the game state. Mathematically, it can be written as: 

\begin{equation}
 DNTS(s_N) = \min_{s* \in \mathcal{S}} ||\mathcal{F}(s*) - \mathcal{F}(s_N)||_2^2 
\end{equation}

where $s_N$ is the current state, $\mathcal{F}(\cdot)$ is the feature embedding of the q-network, and $\mathcal{S}$ is the collection of all game states in the training buffer. Together, these metrics can be combined to give the user a sense of how well the agent is estimating the current state value relative to how similar the game states are to what it has seen in training. We present the information in a 2-D graph that is "traced-out" along the course of gameplay. This is shown in Fig. \ref{UI}. 

\subsection{Creating Similar Scenarios}
\begin{figure*}
\includegraphics[width=1\linewidth]{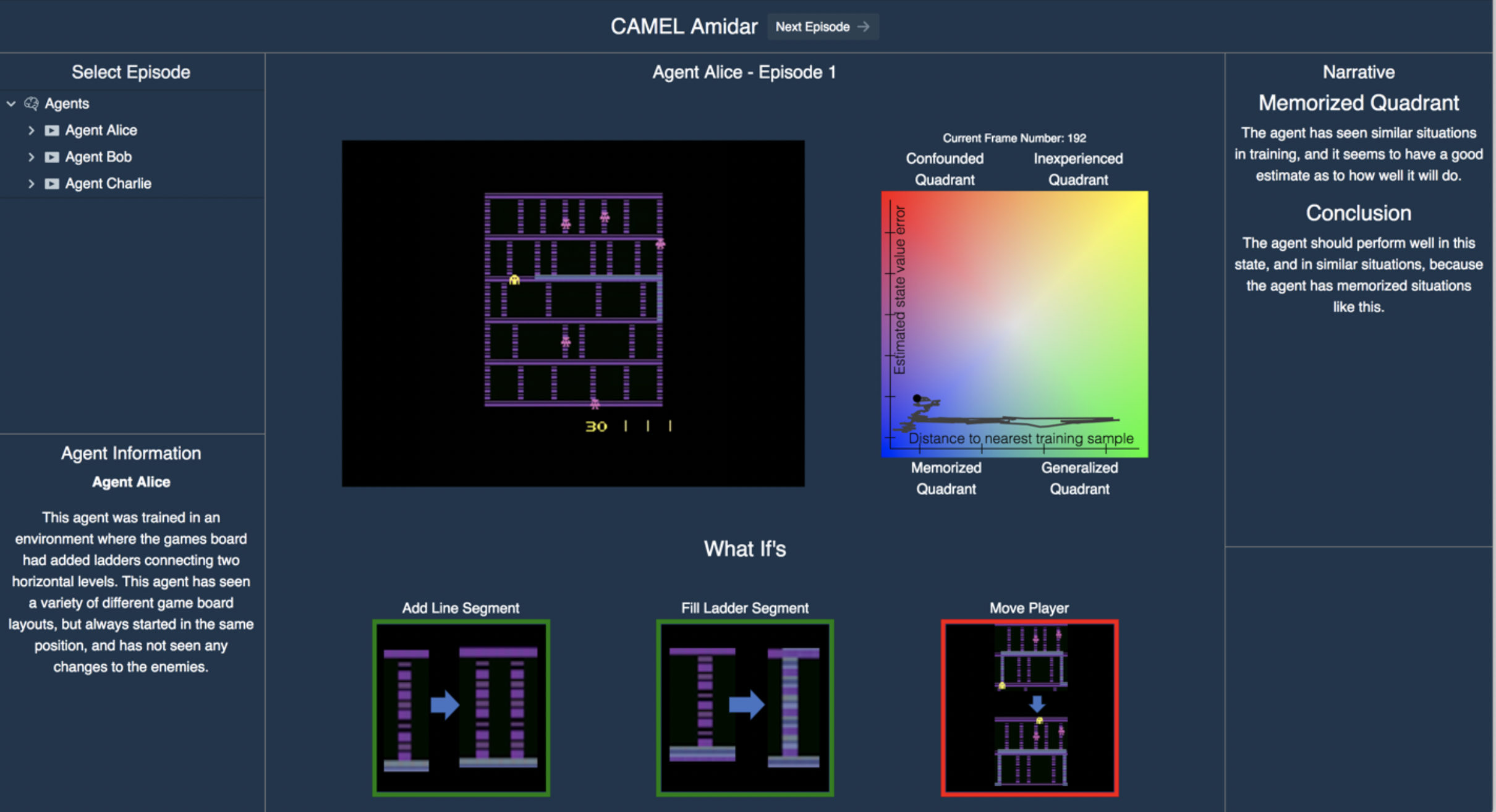}
\caption{Our explanation interface for Amidar.}
\centering
\label{UI}
\end{figure*}

A primary motivation for conveying trustworthiness of autonomous agents was observing that agents were in fact very brittle with respect to their input differing from their training environments. A consequence of this was that a user of the system would be left with an erroneous mental model of the AI. E.g., in our Amidar environment, the agent appeared to allude enemies with sophisticated strategies such as "hiding" in a position waiting for the enemies to pass, or dodging into nearby levels. This can result in an observer hallucinating that the agent has learned complex strategic tactics. To mitigate this problem, we provide the user with an analysis of how the agent would perform on \emph{intervened} game states. That is, on game states that are similar and still semantically meaningful (i.e., the game rules/appearance is preserved), but differ in some way. To show an analysis of how the agents will respond in a similar (but different) environment, we chose the following interventions: adding vertical line segments between horizontal levels, filling in line segments, and moving the agent location on the board.(see Section \ref{UI_section}, and Fig. \ref{UI}).

We perform interventions by creating an emulator that creates Amidar game state. These high-level interventions to the game were enabled by having access to the Atari Toolbox \cite{tosch2019toybox}. For consistency, the same renderer that generated the game environment during training, also rendered the manipulated environments. 

\begin{figure*}
\includegraphics[width=1\linewidth]{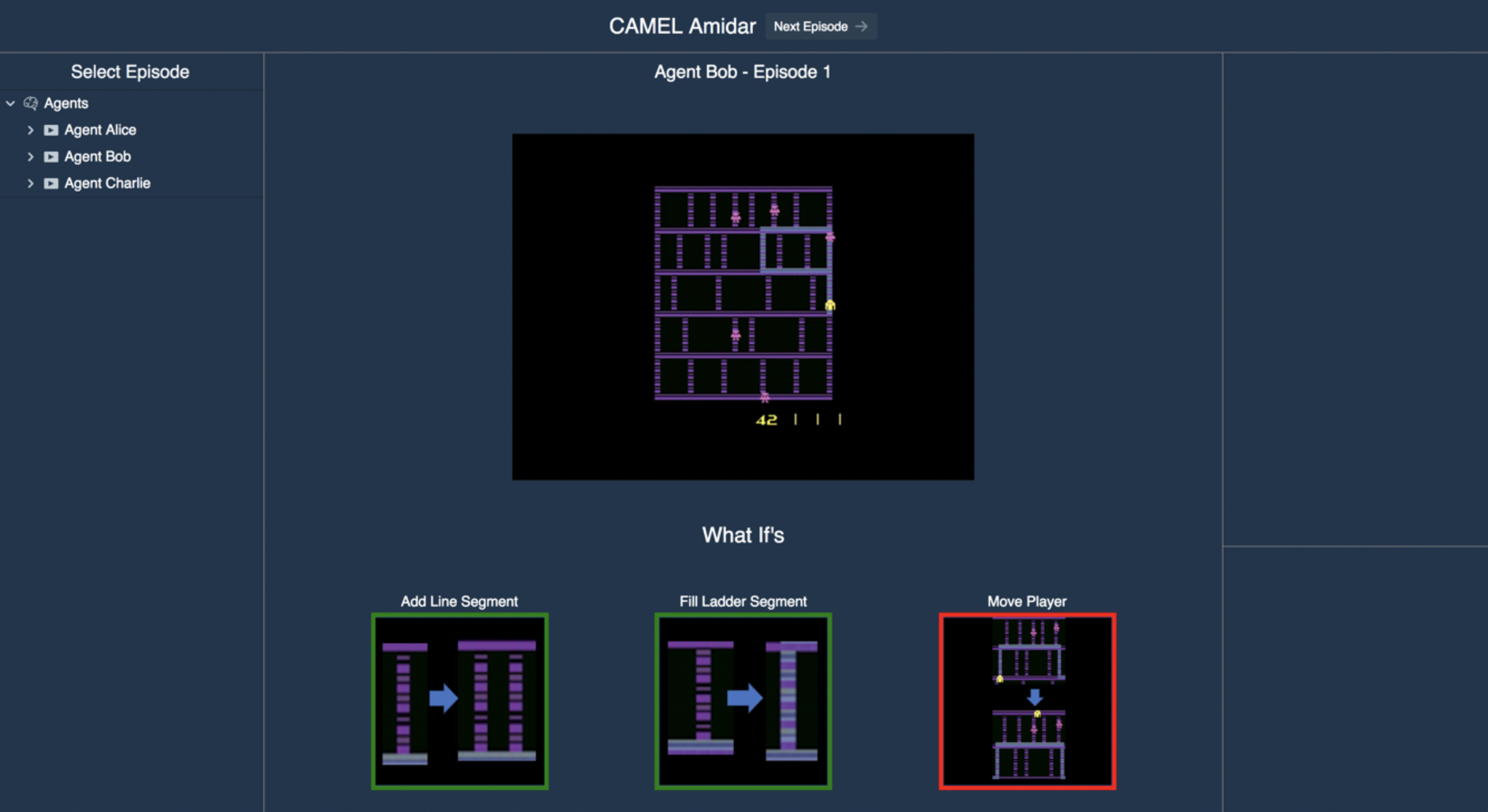}
\caption{The user-interface for Amidar without explanation.}
\centering
\label{UI_no}
\end{figure*}

\subsection{The User Interface} \label{UI_section}
A user-interface (UI) can be an effective tool in establishing user trust in a software system \cite{klein1994framework, carlson2014identifying}. For our XAI system, we created a user-interface (UI) with six major components (see Fig. \ref{UI}). 

\begin{enumerate}
\item \textbf{Agent selection} For our initial framework, we let this be a selection of 3 agents, each trained in an Amidar environment with different game-board characteristics. The first agent was trained on an unmodified Amidar game-board. The second agent was trained on an Amidar board which contains random ladders connecting two horizontal levels (traversable by the agent and enemies). The third agent was trained on an Amidar game-board where the player had a randomized starting position. This is to verify our approach is applicable to different autonomous agents for Amidar.  
\item \textbf{Agent description} A brief description of how the agent was trained. E.g., agent Alice has been trained in a scenario where some ladders have been randomly added in its environment. 
\item \textbf{Live game stream} After the user selects an agent, the live video of the game progressing is displayed. 
\item \textbf{Agent performance characterization} The 2D-figure traces out the agent's VEE and DNTS metrics (see Section \ref{metrics} along the course of the game. 
\item \textbf{Narrative explanation} Narrative statements which provide a textual interpretation of the VEE and DNTS; 
\item \textbf{"What If" scenarios} Indicate performance on how the agent \emph{would} play if particular interventions were performed. For each intervention, a green border indicates the agent would have received over 75\% of the mean reward for the agent playing games with no interventions; a red border indicates the agent would have scored below 75\%. The first intervention type is "Add Line Segment", this intervention consists of adding a vertical line segment connecting two horizontal levels in a game-board. The second intervention is "Fill Ladder Segment", this intervention consists of filling or "painting" a line segment (this occurs when an agent traverses any point on the game-board). The third intervention is "Move Player", which consists of transporting the agent to another legal position on the game-board (player is never placed directly in a place where they will be immediately killed). 
\end{enumerate}

\section{User Study} \label{experiment}
\subsection{Evaluation Design}

The evaluation design includes two tasks: 
\begin{itemize} 
\item Acceptance Testing. Participants watch 3 agents perform in 12 video sequences either with or without explanations, and afterwards complete a questionnaire on system acceptance.
\item System Prediction. Participants view 36 (freeze-frame) Amidar game states, and are asked to  predict system performance for each of the 3 agents.
\end{itemize}

\begin{figure}
\includegraphics[width=1.0\linewidth]{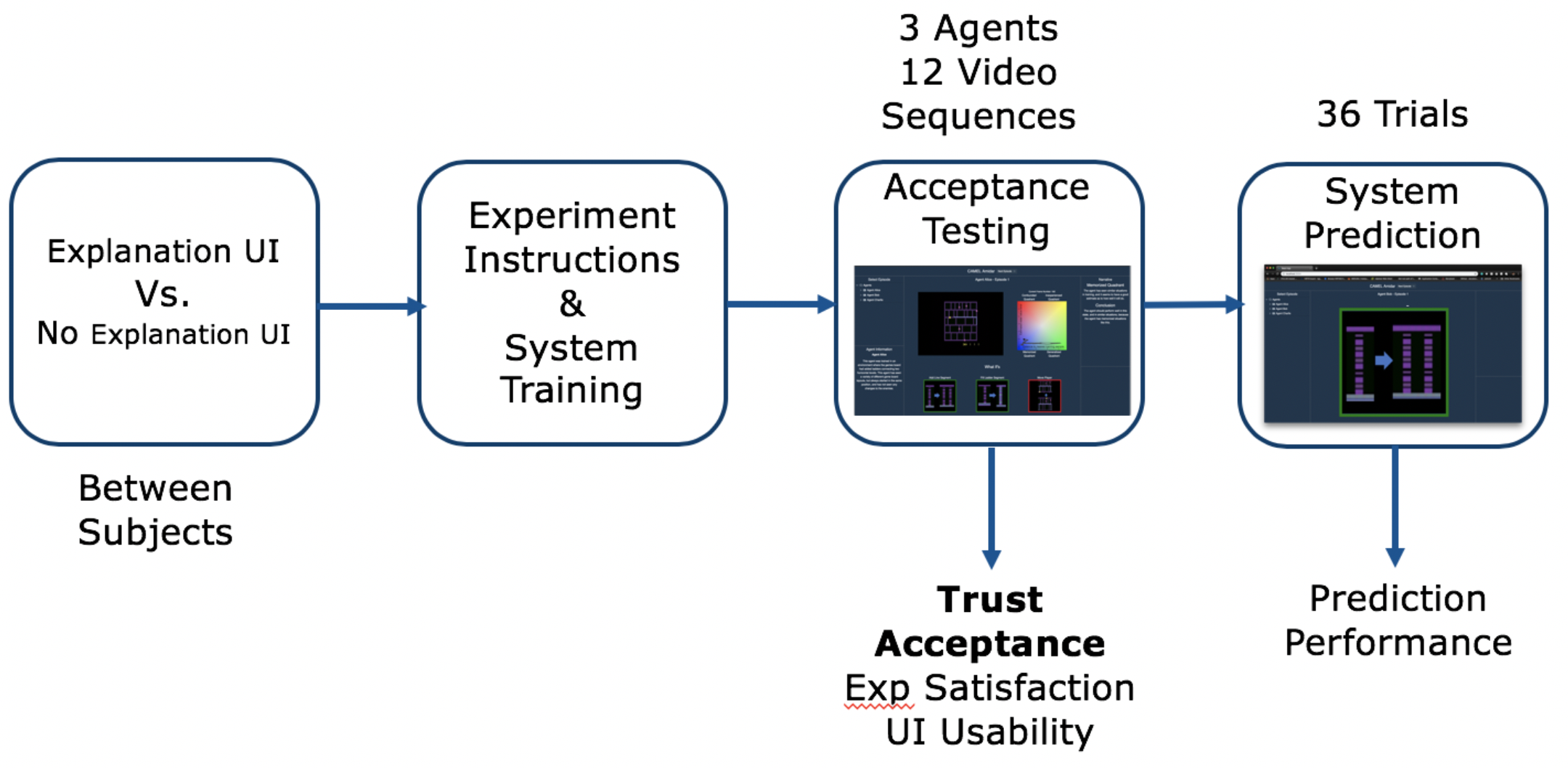}
\caption{Workflow for our experiment.}
\centering
\label{workflow}
\end{figure}

An overview for the participant workflow design for the evaluation can be seen in Fig. \ref{workflow}. Participants were run in groups of 4, unless a scheduled participant did not arrive for the session, and all participants in a group participated in the same condition—either explanation or no explanation. Participants were informed that they would be viewing an automated agent play a simple video game, and that their task was to view the system output over 36 sequences, and then assess whether they thought the automated agent should or should not be accepted as useful. The participants in the Explanation UI condition were given an additional training session of approximately 15 minutes during which the main elements of the Explanation Interface were described using numerous examples, and participants were able to ask questions about interface functionality. Because of the simplicity of the interface in the No Explanation UI condition, the training for those participants was less than five minutes, but participants also had the opportunity to ask any questions about interface functionality before the study began. Fig \ref{UI} shows the Explanation Interface for the AMIDAR game playing agent and Fig. \ref{UI_no} shows the control interface.

\subsection{Experiment Results}
After the acceptance testing task, all participants were given both trust and user acceptance questionnaires. In addition, explanation satisfaction (Hoffman, Mueller, Klein, and Litman, 2018) and usability/usefulness questionnaires were administered to participants in the Explanation Interface condition. All questionnaires used a five point Likert scale, and only included positive statements about either the autonomous system or the XAI interface so that higher scores always indicated more positive user assessments of the autonomous system, and the XAI interface.

Below are the questions answered by the subjects (note the Liker scale was 1 through 5, with 1 being "Strongly Disagree", and 5 being "Strongly Agree"):\\ \\
\textbf{Trust Questions }
\begin{enumerate}
\item I understand how the Automated Game Player works – its goals, actions and output.
\item The Automated Game Player is reliable. It plays the game in a consistent manner. 
\item It is easy to follow what the Automated Game Player does.
\item I am able to predict how the Automated Game Player will play the game (Its actions are predictable.)
\item I am able to judge when I should trust and not trust the Automated Game Player.
\item I am confident in the Automated Game Player.  I feel that it performs well.
\item The Automated Game Player UI will have a positive effect on my performance when it comes to understanding Agent capabilities.
\item I can trust the Automated Game Player UI.
\end{enumerate}

\textbf{User Acceptance Questions}
\begin{enumerate}
\item The Automated Game Player User Interface would be a useful aid in understanding how an Automated player will perform in different game conditions
\item The Automated Game Player User Interface enables a person to understand how the automated player will perform in different game conditions more quickly and efficiently.
\item The Automated Game Player User Interface enables a person to understand how the automated player will perform in different game conditions better than they could on their own.
\item The Automated Game Player User Interface is easy to use
\end{enumerate}

\begin{figure*}[t]
\includegraphics[width=0.75\linewidth]{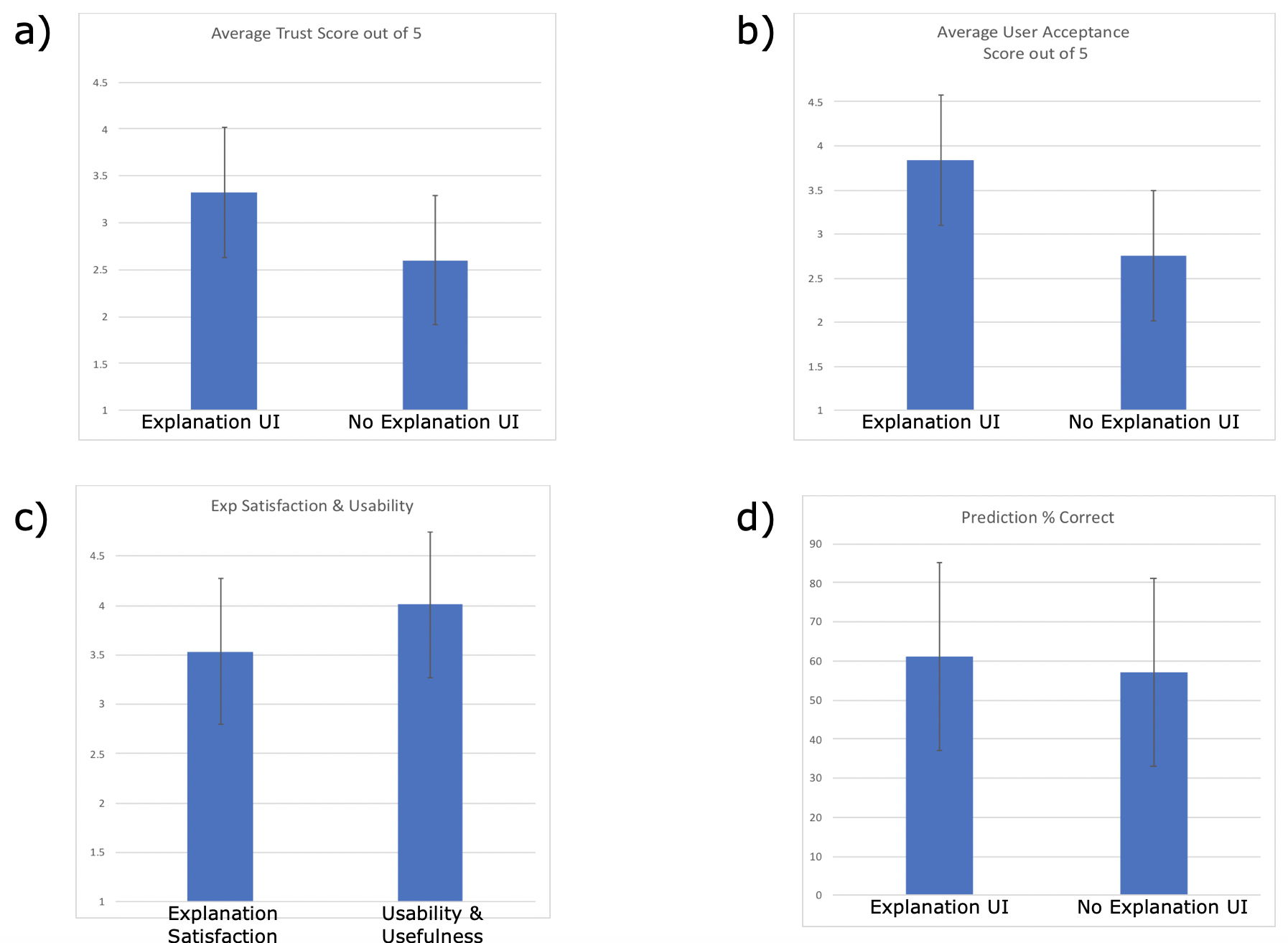}
\caption{Workflow for our experiment.}
\centering
\label{results}
\end{figure*}

Fig. \ref{results} a) shows a greater average Trust score for the Explanation UI condition (3.32) compared to the No Explanation UI condition (2.6). A statistical hypothesis test indicated this difference was not significant (t (df 10) = 1.71, p = .058). 
Fig. \ref{results} b) shows a greater average User Acceptance score for the Explanation UI condition (4.02) compared to the No Explanation UI Condition (2.67). A statistical hypothesis test indicated this difference was significant (t (df 10) = 2.72, p = .03). 
Based on the 5-point Likert scale, an average user response should be 3, and therefore the results in Fig. \ref{results} c) indicate that the average Usability/Usefulness (4.01) and Explanation Satisfaction scores (3.53) are both higher than the neutral scale value. 
We hypothesized that the Explanation Interface would provide users with a richer mental model of the AI classifier, which would lead to a higher percent correct on the prediction task. However, the results in Fig. \ref{results} d) show only slightly greater Prediction 

\section{Conclusions and Future Work} \label{conclusion}
The results of the evaluation confirm that our XAI system did indeed increase user trust and acceptance of the AI system. This finding is both supported from qualitative questionnaires, as well as statistical results from quantitative questions. Given the brittle nature of the black box AI systems, this result is encouraging that user's value on information as to the trustworthiness of AI systems. However, the results are not exceedingly positive, with improvements barely outside of a single standard deviation. Specifically, the perception of prediction correctness did not significantly increase with explanations. This implies the users perhaps did not learn a significantly better mental model for the autonomous system. 

Several research directions naturally follow from this work. First, we would like to consider a more complex domain, necessitating a more sophisticated deep RL policy network. For example, in many critical tasks, AIs will not act autonomously, but act as trusted partners in a human-machine collaboration. Providing an XAI framework for this hybrid task to check for increase efficiency would be interesting. Additionally, the manner in which the VEE was calculated(see \ref{metrics}), could be improved. E.g., we could use the cumulative error up until the current point in time (rather than "artificially" rolling out the episode until completion). A study as to the effectiveness of this approximation would be of interest.

\subsubsection*{Acknowledgments}

The authors would like to thank Amanda DiFiore for her valuable contributions during data collection. This research was developed with funding from the Defense Advanced Research Projects Agency (DARPA). This material is based on research sponsored by the Air Force Research Lab (AFRL) under agreement number FA8750-17-C-0118. The U.S. Government is authorized to reproduce and distribute reprints for governmental purposes notwithstanding any copyright notation thereon. The views, opinions and/or findings expressed are those of the author and should not be interpreted as representing the official views or policies of the Department of Defense or the U.S. Government.

\medskip

\small

\bibliographystyle{ACM-Reference-Format}
\bibliography{sample-base}
\end{document}